\documentclass[10pt,twocolumn,letterpaper]{article}

\usepackage{cvpr}              

\usepackage[dvipsnames]{xcolor}

\definecolor{cvprblue}{rgb}{0.21,0.49,0.74}
\usepackage[pagebackref,breaklinks,colorlinks,citecolor=cvprblue]{hyperref}

\usepackage{float}
\usepackage{multirow}
\usepackage{graphicx}
\usepackage{colortbl}
\usepackage{utfsym}
\usepackage{subcaption}
\usepackage{tabularx}
\usepackage{caption}
\usepackage{wrapfig}
\usepackage{graphicx}
\usepackage{multirow}
\usepackage{colortbl}
\usepackage{booktabs}
\usepackage{amssymb}
\usepackage{bbding}
\usepackage{pifont}
\usepackage{wasysym}
\usepackage{fontawesome}
\usepackage[export]{adjustbox}

\usepackage{pifont}

\definecolor{sh_blue}{rgb}{0,0.60,0.93}
\definecolor{sh_gray2}{rgb}{1,0.89,0.75}
\definecolor{color3}{rgb}{0.95,0.95,0.95}
\definecolor{mygray}{gray}{.9}
\definecolor{bluegreen}{rgb}{0.44, 0.64, 0.77}
\definecolor{gray_venue}{rgb}{0.53,0.52,0.52}
\definecolor{color5}{rgb}{1,0.96,0.88}

\newlength{\Oldarrayrulewidth}

\usepackage[accsupp]{axessibility}  

\def\paperID{***} 
\def\confName{CVPR}
\def\confYear{2024}

\title{Teaching Tailored to Talent: \\ 
Adverse Weather Restoration via Prompt Pool and Depth-Anything Constraint}

\author{
Sixiang Chen$^{1}$ \quad Tian Ye$^{1}$ \quad Kai Zhang$^{1}$ \quad Zhaohu Xing$^{1}$ \quad Yunlong Lin$^{2}$ \quad Lei Zhu$^{1,3}$\thanks{Lei Zhu (leizhu@ust.hk) is the corresponding author.}\\ \vspace{-0.5mm}
  $^{1}$The Hong Kong University of Science and Technology (Guangzhou)\quad
  $^{2}$Xiamen University\quad \\ \vspace{-0.5mm}
  $^{3}$The Hong Kong University of Science and Technology\\
{\tt\small Project page: \url{https://ephemeral182.github.io/T3-DiffWeather}}}

\begin{document}






\maketitle

\begin{abstract}
Recent advancements in adverse weather restoration have shown potential, yet the unpredictable and varied combinations of weather degradations in the real world pose significant challenges. Previous methods typically struggle with dynamically handling intricate degradation combinations and carrying on background reconstruction precisely, leading to performance and generalization limitations. Drawing inspiration from prompt learning and the "\underline{\textbf{T}}eaching \underline{\textbf{T}}ailored to \underline{\textbf{T}}alent" concept, we introduce a novel pipeline, \textbf{T$^{3}$-DiffWeather}. Specifically, we employ a prompt pool that allows the network to autonomously combine sub-prompts to construct weather-prompts, harnessing the necessary attributes to adaptively tackle unforeseen weather input. Moreover, from a scene modeling perspective, we incorporate general prompts constrained by Depth-Anything feature to provide the scene-specific condition for the diffusion process. Furthermore, by incorporating contrastive prompt loss, we ensures distinctive representations for both types of prompts by a mutual pushing strategy. Experimental results demonstrate that our method achieves state-of-the-art performance across various synthetic and real-world datasets, markedly outperforming existing diffusion techniques in terms of computational efficiency.

\end{abstract}
\vspace{-0.6cm}

\section{Introduction}
\vspace{-0.1cm}
\label{sec:intro}
With the growth of the community, image restoration in adverse weather conditions has become increasingly significant~\cite{IDT,chen2023learning,chen2023sparse,song2022vision,chen2020jstasr,hdcwnet,jin2022structure}. To meet practical demands effectively, research is increasingly focusing on the all-in-one removal of multiple weather degradations~\cite{yang2024genuine,valanarasu2022transweather,chen2022learning,weatherdiff,allinone} as a primary objective.

Contrasting with restoration tasks targeted at specific weather conditions, multi-weather restoration involves a composition of various weather phenomena.
%
Early developments employed methods such as neural architecture search~\cite{allinone} or distillation~\cite{chen2022learning} to combine models tailored to individual tasks, but these approaches are complicated and cumbersome. 
Moreover, several methods~\cite{ye2023adverse,valanarasu2022transweather} have attempted to employ a codebook as a reliable prior for guiding image restoration or utilized shared learnable queries to adapt different weather degradations. However, such paradigms are not aware of the differences and similarities 

\begin{figure}[t]
\vspace{-0.2cm}
    \centering
    \setlength{\abovecaptionskip}{0.2cm} 
    \setlength{\belowcaptionskip}{-0.2cm}
    \includegraphics[width=8cm]{ 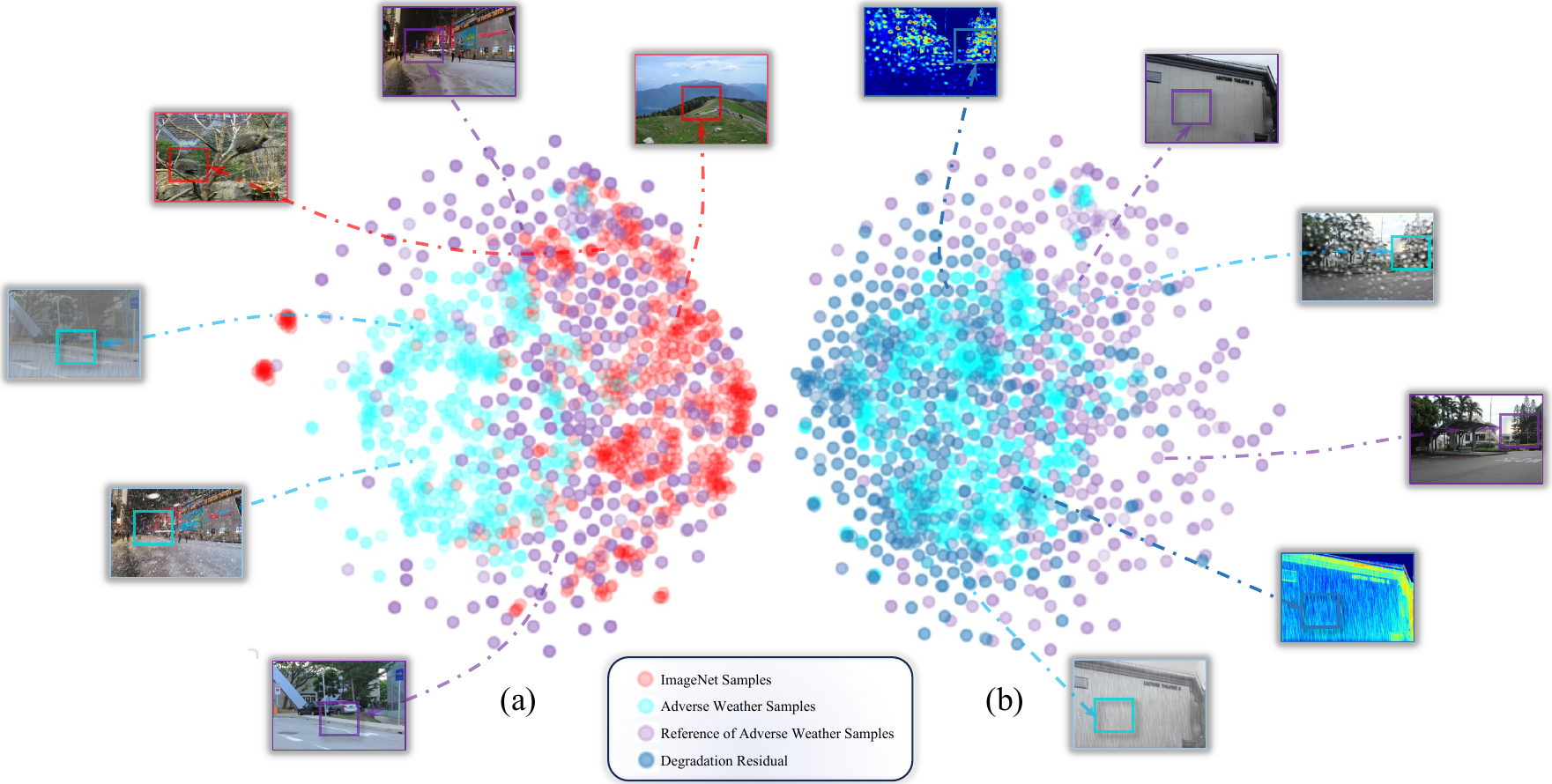}
    \caption{\small{\textbf{t-SNE visualization of different feature distributions.} (a). Scenes with different contents also have significant commonalities compared to degradations. And there are some differences and commonalities between degradations and degradations. (b). Degradation residuals can represent degradations to a certain extent and be distinguished from the background. }}
    \label{motivation}
    \vspace{-0.4cm}
\end{figure}

\noindent between degradations. Recently, the WGWS framework~\cite{zhu2023learning} analyzed the weather-general and weather-specific features and performed targeted parameter learning for such characteristics. Yet, its two-stage design and the need for customized modifications within different architectures remain intricate. 
In addition, unlike dealing with a single weather, adverse weather will cause unseen and unpredictable degradation combinations in the real world, which poses a challenge to handling degradations adaptively.
Hence, \textbf{\textit{1. How to effectively and flexibly model complicated and unpredictable weather combinations in the real world remains an open question}}.
 
Furthermore, benefiting from recent advancements in diffusion models~\cite{ddpm}, there is the first diffusion-based method—WeatherDiffusion~\cite{weatherdiff}—which has demonstrated the superiority of generative paradigms over regression models in reconstructing clean backgrounds from adverse weather images. 
However, its shortcomings are evident: the original degraded image as a condition does not adequately guide the reconstruction from noise to clean images, and requires a certain number of steps for sampling. For diffusion models, discovering sufficiently informative conditions is essential for high-quality image reconstruction~\cite{guo2023shadowdiffusion,lin2023diffbir,xia2023diffir,jin2024des3}. Therefore, \textbf{\textit{2. Designing a condition that equips rich information on adverse weather samples is critical for diffusion process.}}

To address the aforementioned challenges, we introduce a novel paradigm. Inspired by prompt learning~\cite{vpt,coop}, we claim that prompts can be employed to craft a comprehensive condition. \textbf{i).} Unlike the recent paradigm~\cite{potlapalli2023promptir} that utilized a shared set of learnable prompts to adapt to varying degraded images, we design a prompt pool. This pool can fully exploit the differences and similarities among weather degradations, thereby offering the network a wider range of options. Specifically, we enable the network to autonomously construct the necessary weather information based on the input degradation residual and freely assemble a specific set of weather-prompts for unpredictable phenomena.
\textbf{ii).} We observe that the scene features behind adverse weather often share commonalities (see Fig.\ref{motivation} (a)), inspiring us to devise general prompts specifically for background modeling. Drawing inspiration from the recent breakthroughs in Depth-Anything~\cite{yang2024depth}, we note this model exhibits exceptional robustness when handling extreme samples. This observation leads us to extract critical features from Depth-Anything, utilizing them to direct our general prompts.
\textbf{iii).} Additionally, we introduce a compact contrastive prompt loss to further regulate two types of prompts, and integrate them seamlessly into the diffusion process. At the core of our approach, a "\underline{\textbf{T}}eaching \underline{\textbf{T}}ailored to \underline{\textbf{T}}alent" paradigm is resembled , adeptly guiding distinct weather combinations and their respective scene backgrounds. Building upon this foundation, we introduce a novel diffusion-based architecture: \textbf{\textit{T$^{3}$-DiffWeather}}. 
It achieves SOTA performance across various adverse weather benchmarks while requiring only a tenth of the sampling steps compared to the latest WeatherDiffusion~\cite{weatherdiff}.

As an overview, the contributions of our work are summarized as follows:
\vspace{-0.15cm}
\begin{itemize}
    \item We introduce a novel prompt pool. Capitalizing on the similarities and differences among various weather conditions, proposed prompt pool empowers the network to autonomously combine sub-prompts, effectively constructing diverse weather-prompts to enhance representation for complicated weather degradations.
    \item Inspired by the shared attributes within the scenes of degraded samples, we have crafted general prompts specifically tailored for background understanding. For the first time, we propose to utilize the robust features of Depth-Anything~\cite{yang2024depth} as a constraint for general prompts.
    \item We incorporate a compact contrastive prompt loss to further boost the prompt representation of two designs. Overall, our diffusion-based architecture, T$^{3}$-DiffWeather, achieves SOTA performance on multi-weather restoration benchmarks with significantly fewer inference steps.
\end{itemize}
\vspace{-0.2cm}

\section{Related Works}
\vspace{-0.1cm}
\subsection{Adverse Weather Restoration}
\vspace{-0.1cm}
\noindent{\textbf{Dehazing}}: The field of single-image dehazing has witnessed remarkable advancements in recent years~\cite{mbdehaze,wu2021contrastive,dehazenet,song2022vision,griddehazenet,ye2021perceiving,liu2023nighthazeformer}. 
DehazeFormer~\cite{song2022vision} adopted a transformer-based approach, tackling the complicated hazy images through distinct window processing. Meanwhile, the MB-TaylorFormer~\cite{mbdehaze} leveraged a linear Transformer architecture grounded in Taylor series expansion to clarify hazy scenes effectively.

\noindent{\textbf{Deraining}}: The progress in single-image rain removal is steadily increasing, including rain streaks~\cite{prenet,JORDER,zhu2018bidirectional,chen2023learning} and raindrops~\cite{qian2018attengan,quaninonego,chen2023sparse,IDT,wu2023mask}. IDT~\cite{IDT} developed a transformer-based technique that combines window-based and spatial transformers to enhance the precision of rain streak modeling. UDR-S2Former~\cite{chen2023sparse} leveraged uncertainty to refine the sparse ViT model for improved performance of raindrop removal.

\noindent{\textbf{Desnowing}}: Unlike dehazing and deraining, single-image snow removal presents a greater challenge~\cite{liu2018desnownet,chen2020jstasr,cheng2022snow,chen2022snowformer,chen2022msp,chen2023snow,chen2023uncertainty,ye2022towards}. 
JSTASR~\cite{chen2020jstasr} introduced a framework capable of addressing both haze and snow removal simultaneously. MSP-Former~\cite{chen2022msp} was the inaugural attempt at a single-image snow removal network utilizing a transformer architecture. Nevertheless, similar to haze and rain removal, these innovative approaches still grapple with limitations when confronted with other variants of extreme weathers.

\noindent{\textbf{Multi-Weather Restoration}}: Adverse weather restoration endeavors to develop a consolidated network to adeptly address weather-induced image degradations~\cite{allinone,valanarasu2022transweather,chen2022learning,weatherdiff,ye2023adverse,zhu2023learning,yang2023video}. The pioneering work in this domain was the All-in-One~\cite{allinone}, the extensive parameterization due to NAS may make it impractical for real-world deployment. TransWeather~\cite{valanarasu2022transweather} introduced a weather-type decoder capable of interpreting diverse weather degradations, yet its fixed queries cannot explicitly consider the degradations of different weather and lacks background-level modeling. 
WeatherDiffusion~\cite{weatherdiff} presented a diffusion-based method that harnessed the capabilities of diffusion models for weather removal, achieving SOTA results across various benchmarks. Nonetheless, the slow inference speed and the absence of precise prompt conditions may hinder its widespread practical application.
\vspace{-0.1cm}

\subsection{Conditional Diffusion Models}
\vspace{-0.1cm}
Recent advancements in denoising diffusion probabilistic models (DDPM)~\cite{diffuision} have captured intricate distributions with accuracy that exceeds other generative frameworks, including GANs. To further enhance the precision and realism of the generated outputs, diffusion models often incorporated additional conditioning or guidance mechanisms, as evidenced by recent studies~\cite{guo2024versat2i,cao2023difffashion,cao2023image,jiang2024back}.
In the field of image restoration, the prevailing approach involves feeding networks with concatenated degraded inputs to yield outputs of high fidelity quality~\cite{weatherdiff,sr3,li2022srdiff,jin2024des3,ye2024learning} compared with traditional regressive models~\cite{jin2021dc,jin2022unsupervised,jin2023enhancing,wang2023masked,ren2024rethinking,jiang2023five,fang2024spiking}. To further refine the denoising process, networks often incorporated single task-specific prompts such as masks or textual information~\cite{guo2023shadowdiffusion,yang2023pixel} as embedded guidance. However, the recent WeatherDiffusion~\cite{weatherdiff} typically used degraded images as the sole condition,  which may result in performance limitations for addressing all-in-one restoration tasks.
\vspace{-0.1cm}
\subsection{Prompt Learning}
\vspace{-0.1cm}
Prompt learning has been increasingly applied to computer vision~\cite{song2024moviechat,chai2023stablevideo,zhao2023see,ouyang2023chasing,song2024moviechat+,chai2022deep}. This technique involved the insertion of task-specific prompt tokens prior to the input, equipping pre-trained models with the necessary knowledge to perform new tasks without extensive fine-tuning. The approach of Context Optimization (CoOp)\cite{coop} leveraged this for the CLIP model\cite{clip}, refining prompts in a continuous space through backpropagation. Conditional Context Optimization (CoCoOp)\cite{cocoop} innovated further by producing input-dependent prompt residuals to enhance generalization ability. 
For low-level tasks, PromptIR~\cite{potlapalli2023promptir} introduced a learnable prompt module that generates shared prompts responsive to various degradation types.  Additionally, recent researches utilized text prompts to guide image restoration networks~\cite{lin2023diffbir,chen2023image,sun2024coser}. However, the use of sparse text embeddings could lead to performance limitations and increase complexity due to the necessity for additional multi-modal models.
\vspace{-0.1cm}
\section{Preliminaries}
\vspace{-0.1cm}
\noindent{\textbf{Diffusion Models.}}
Diffusion models (DMs)~\cite{ddpm,nichol2021improved} infuse training data with Gaussian noise and then recover the original data through the inversion of this noise.
Initially, DMs implement a diffusion algorithm that incrementally converts a starting image $\boldsymbol{x}_{0}$ into a noise distribution $\boldsymbol{x}_{T}\sim \mathcal{N}(0,1)$ across $T$ steps. Each step of this transformation is described by the equation:
\begin{equation}\label{eq1}
q\left(\boldsymbol{x}_t | \boldsymbol{x}_{t-1}\right) = \mathcal{N}\left(\boldsymbol{x}_t ; \sqrt{1-\beta_t} \boldsymbol{x}_{t-1}, \beta_t \boldsymbol{I}\right),
\end{equation}
with $\boldsymbol{x}_t$ representing the image with noise at timestep $t$, $\beta_t$ as a predetermined scale parameter, and $\mathcal{N}$ signifying the Gaussian distribution.
Defining ${\alpha}_{t}=1-{\beta}_{t}$ and $\bar{\alpha}_t=\prod_{i=0}^t {\alpha}_{i}$ allows us to simplify Eq.\ref{eq1} as:
\begin{equation}\label{eq2}
q\left(\boldsymbol{x}_t | \boldsymbol{x}_0\right) = \mathcal{N}\left(\boldsymbol{x}_t ; \sqrt{\bar{\alpha}_t} \boldsymbol{x}_0,(1-\bar{\alpha}_t) \boldsymbol{I}\right).
\end{equation}

During the inference phase, DMs commence by generating a Gaussian noise map $\boldsymbol{x}_{T}$ and then progressively apply a denoising process until reaching a high-fidelity result $\boldsymbol{x}_{0}$:
\begin{equation}
p\left(\boldsymbol{x}_{t-1} | \boldsymbol{x}_t, \boldsymbol{x}_0\right) = \mathcal{N}\left(\boldsymbol{x}_{t-1} ; \boldsymbol{\mu}_t(\boldsymbol{x}_t, \boldsymbol{x}_0), \sigma_t^2 \boldsymbol{I}\right),
\end{equation}
where the mean value $\boldsymbol{\mu}_t(\boldsymbol{x}_t, \boldsymbol{x}_0)=\frac{1}{\sqrt{\alpha_t}}\left(\boldsymbol{x}_t - \boldsymbol{\epsilon} \frac{1-\alpha_t}{\sqrt{1-\bar{\alpha}_t}}\right)$ and the variance $\sigma_t^2=\frac{1-\bar{\alpha}_{t-1}}{1-\bar{\alpha}_t} \beta_t$. The noise estimate $\boldsymbol{\epsilon}$ is optimized $\boldsymbol{\epsilon}_\theta(\boldsymbol{x}_t, t)$ through training as follows: with a clean image $\boldsymbol{x}_0$, DMs select a random timestep $t$ and noise $\boldsymbol{\epsilon} \sim \mathcal{N}(0, \boldsymbol{I})$ to produce the noisy images $\boldsymbol{x}_t$ as per Eq.\ref{eq2}. DMs then optimize the model parameters $\theta$ of $\boldsymbol{\epsilon}_\theta$ in accordance with~\cite{ddpm}:
\begin{equation}\label{eq4}
\mathcal{L}_{diff} = \mathbb{E}\left\|\boldsymbol{\epsilon}-\boldsymbol{\epsilon}_{\boldsymbol{\theta}}\left(\sqrt{\bar{\alpha}_t} \boldsymbol{x}_0+\boldsymbol{\epsilon} \sqrt{1-\bar{\alpha}_t}, t\right)\right\|_2^2.
\end{equation}

\begin{figure*}[!t]
    \centering
    \setlength{\abovecaptionskip}{0.2cm} 
    \setlength{\belowcaptionskip}{-0.2cm}
    \includegraphics[width=17.5cm]{ 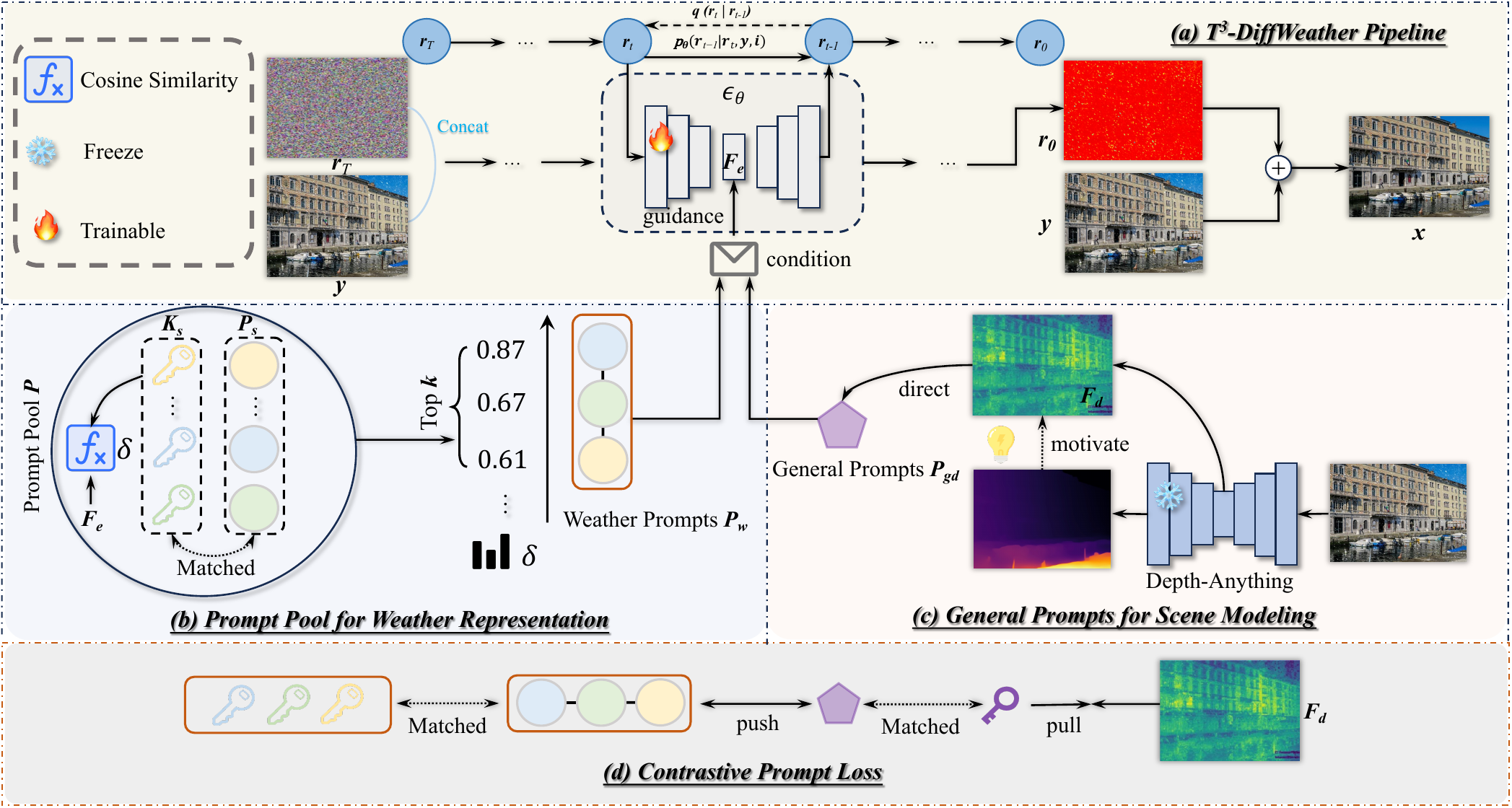}
    \caption{\small{\textbf{The overview of proposed method.} (a) showcases our pipeline, which adopts an innovative strategy focused on learning degradation residual and employs the information-rich condition to guide the diffusion process. (b) illustrates the utilization of our prompt pool, which empowers the network to autonomously select attributes needed to construct adaptive weather-prompts. (c) depicts the general prompts directed by Depth-Anything constraint to supply scene information that aids in reconstructing residuals. (d) shows the contrastive prompt loss, which exerts constraints on prompts driven by two distinct motivations, enhancing their representations. }}
    \label{overview}
    \vspace{-0.2cm}
\end{figure*}

\vspace{-0.1cm}
\section{Methods}
\vspace{-0.1cm}
\vspace{-0.1cm}
\textbf{Novel Pipeline T$^{3}$-DiffWeather}.
For adverse weather restoration, the intricate blend of degradations in the real world poses significant challenges to obtaining clean backgrounds. Consequently, developing a model that can effectively adapt complicated degradation combinations is crucial.
We introduce a novel pipeline, T$^{3}$-DiffWeather, whose key principle is “\underline{\textbf{T}}eaching \underline{\textbf{T}}ailored to \underline{\textbf{T}}alent.” Inspired by prompt learning~\cite{coop,vpt,potlapalli2023promptir}, our design incorporates instance-wise weather-prompts tailored for specific degradations and general prompts for scene information, efficiently exploiting both the disparate and shared attributes present in degraded images. We leverage these prompts as the condition to guide the diffusion process with rich information.

Specifically, leveraging insights from Fig.\ref{motivation}, we observe that degradations exhibit more distinct features compared to the background (as shown in Fig.\ref{motivation}(a)), and degradation residual $\boldsymbol{r}_{d}$ (subtract the degraded image $\boldsymbol{y}$ from the clean image $\boldsymbol{x}$) provides a clearer representation of the degraded image (as illustrated in Fig.\ref{motivation}(b)). We claim that the degradations are a primary factor for the difficulty in restoration. Therefore, we pivot the reconstruction target of the diffusion model toward the degradation residual. The training objective is changed to follows compared with Eq.\ref{eq4}:
\begin{equation}\label{eq:res}
\mathcal{L}_{res}=\mathbb{E}\left\|\boldsymbol{\epsilon}-\boldsymbol{\epsilon}_\theta\left(\sqrt{\bar{\alpha}}(\underbrace{\boldsymbol{x}-\boldsymbol{y}}_{\text {  residual }\boldsymbol{r}_{d}})+\sqrt{1-\bar{\alpha}} \boldsymbol{\epsilon}, \boldsymbol{y}, \boldsymbol{c}\right)\right\|_2^2.
\end{equation}
where $c$ denotes the condition built by weather-prompts $\boldsymbol{\mathcal{P}}_{w}$ and general prompts $\boldsymbol{\mathcal{P}}_{gd}$. 
We simply embed them into the latent layer in the diffusion network through cross-attention, similar to the text embedding in SD~\cite{stable_diff}. This process is naturally efficient and does not take up much computation overhead. Given the feature embedding $\boldsymbol{\mathcal{F}}_{e}\in \mathbb{R}^{H\times W\times D}$ in the latent layer, The formula can be expressed as follows:
\begin{equation}
\begin{split}
\boldsymbol{\mathcal{F}}_{e}^{'} &= \text{softmax}\left(\frac{\boldsymbol{\mathcal{
Q}}(\boldsymbol{\mathcal{F}}_{e}) \boldsymbol{\mathcal{
K}}(\boldsymbol{\mathcal{P}}_{w})^T}{\sqrt{\boldsymbol{\mathcal{D}}}}\right) \boldsymbol{\mathcal{V}}(\boldsymbol{\mathcal{P}}_{w}), \\
\boldsymbol{\hat{\mathcal{F}}_{e}} &= \text{softmax}\left(\frac{\boldsymbol{\mathcal{
Q}}(\boldsymbol{\mathcal{F}}_{e}^{'}) \boldsymbol{\mathcal{
K}}(\boldsymbol{\mathcal{P}}_{gd})^T}{\sqrt{\boldsymbol{\mathcal{D}}}}\right) \boldsymbol{\mathcal{V}}(\boldsymbol{\mathcal{P}}_{gd}),
\end{split}
\end{equation}
where $\boldsymbol{\mathcal{
Q}}(\cdot), \boldsymbol{\mathcal{
K}}(\cdot), \boldsymbol{\mathcal{
V}}(\cdot)$ are the query, key, and value functions. The $\boldsymbol{\hat{\mathcal{F}}_{e}}$ is the output feature embedding.

\vspace{-0.1cm}
\subsection{Prompt Pool for Weather Representation}\label{sec:prompt pool}
\vspace{-0.1cm}

\noindent{\textbf{Motivation I:}} The restoration from adverse weather is often impeded by the complicated and varied combinations of degradations, which can influence network performance. 
%
%
Unlike the substantial domain gap encountered in general restoration between various types of degradations~\cite{chen2023masked,zhang2023crafting,chen2024low}, weather degradations in the real world exhibit some similar attributes, such as haze veiling and low contrast~\cite{zhu2023learning}. Concurrently, degradations specific to distinct conditions manifest unique attributes varying in shape and scale. These differences and similarities inspire us to explicitly leverage degradation characteristics to enhance the representation of degradations.

Leveraging the advancements in prompt learning for image restoration, we posit that the network should adaptively learn the characteristics of degradations and autonomously construct suitable weather-prompts. Consequently, we introduce the design of a prompt pool. This design triggers the network to selectively choose sub-prompts from the pool, crafting a unique weather-prompt tailored to each sample. Such an autonomous construction explicitly takes into account the similarities (shared sub-prompts) and differences (independent sub-prompts) under varying weather conditions.
Specifically, given our prompt pool $\boldsymbol{\mathcal{P}}=\left\{\boldsymbol{\mathcal{P}}_{s}^{i}\right\}_{i=1}^N$ with each sub-prompts $ \boldsymbol{\mathcal{P}}_{s}^{i} \in \mathbb{R}^{L_{s}\times D}$ ($L_{s}$ denotes the length of tokens), where $i$ represents the index of a specific sub-prompts and $N$ is the prompt pool size.
For the input (degradation residual) embedding $\boldsymbol{\mathcal{F}}_{e}$, the weather-prompt construction function $\Psi$ can be defined as:
\begin{equation}
\boldsymbol{\mathcal{P}}_{w} = \Psi(\boldsymbol{\mathcal{P}}, \boldsymbol{\mathcal{F}}_{e}; \Theta),
\end{equation}

\begin{figure}[!t]

    \centering
    \setlength{\abovecaptionskip}{0.2cm} 
    \setlength{\belowcaptionskip}{-0.2cm}
    \includegraphics[width=8cm]{ 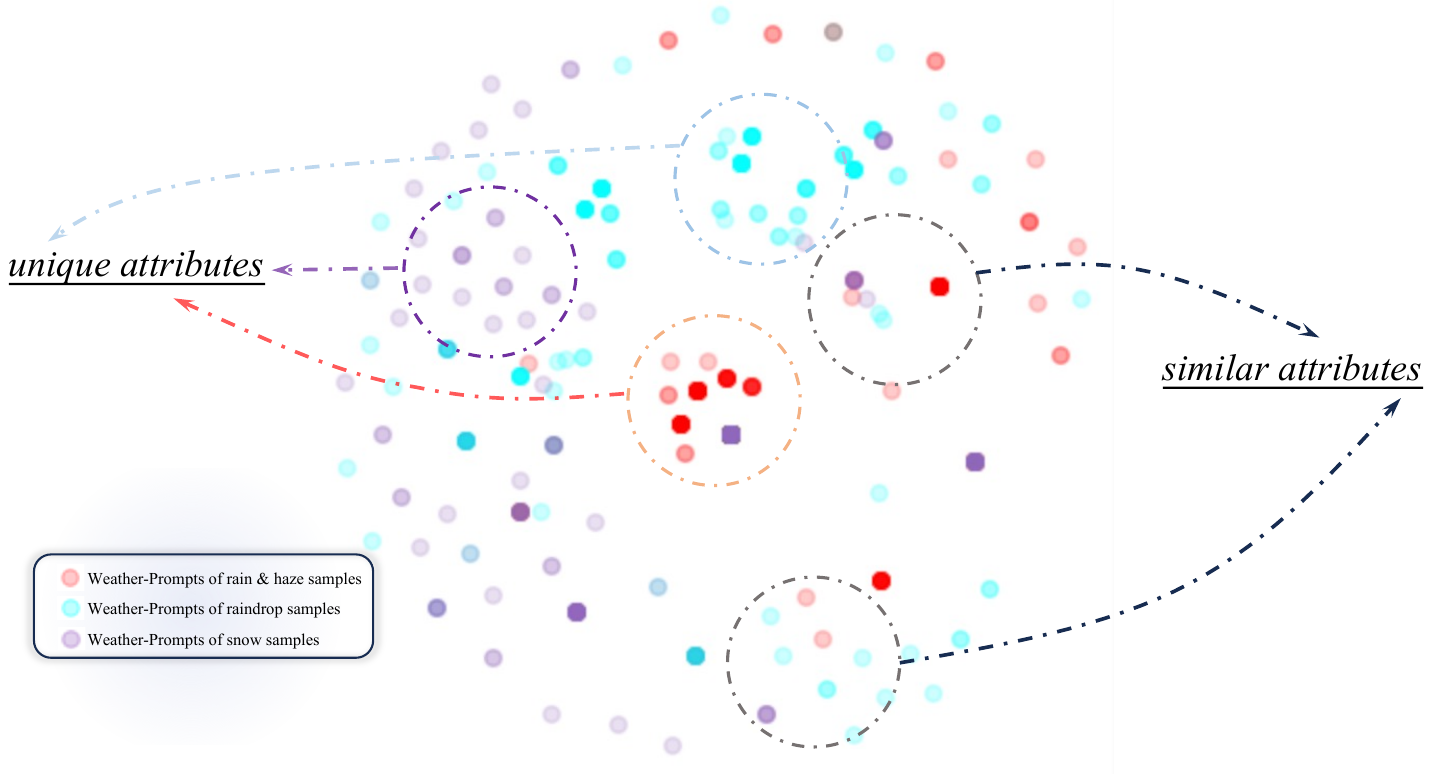}
    \caption{{\textbf{t-SNE visualization of weather-prompts for different weather conditions.}}}
    \label{tsne_vis}
    \vspace{-0.3cm}
\end{figure}

Here, $\Theta$ parameterizes the selection process to optimally align the sub-prompts with the embedded feature (related to degradation) $\boldsymbol{\mathcal{F}}_{e}$.
Motivated by the essence of ViT~\cite{vit}, we utilize the query-key mechanism, which enables the network to select the necessary sub-prompts for the input embedding. Specifically, a learnable key $\boldsymbol{\mathcal{K}}^i_{s} \in \mathbb{R}^{1\times D}$ is designed for each sub-prompts to calculate the correlation with embedding $\boldsymbol{\mathcal{F}}_{e}$ (query) for choose. The formula can be expressed as follows:
\begin{figure}[t]
\vspace{-0.2cm}
    \centering
    \setlength{\abovecaptionskip}{0.2cm} 
    \setlength{\belowcaptionskip}{-0.2cm}
    \includegraphics[width=7.5cm]{ 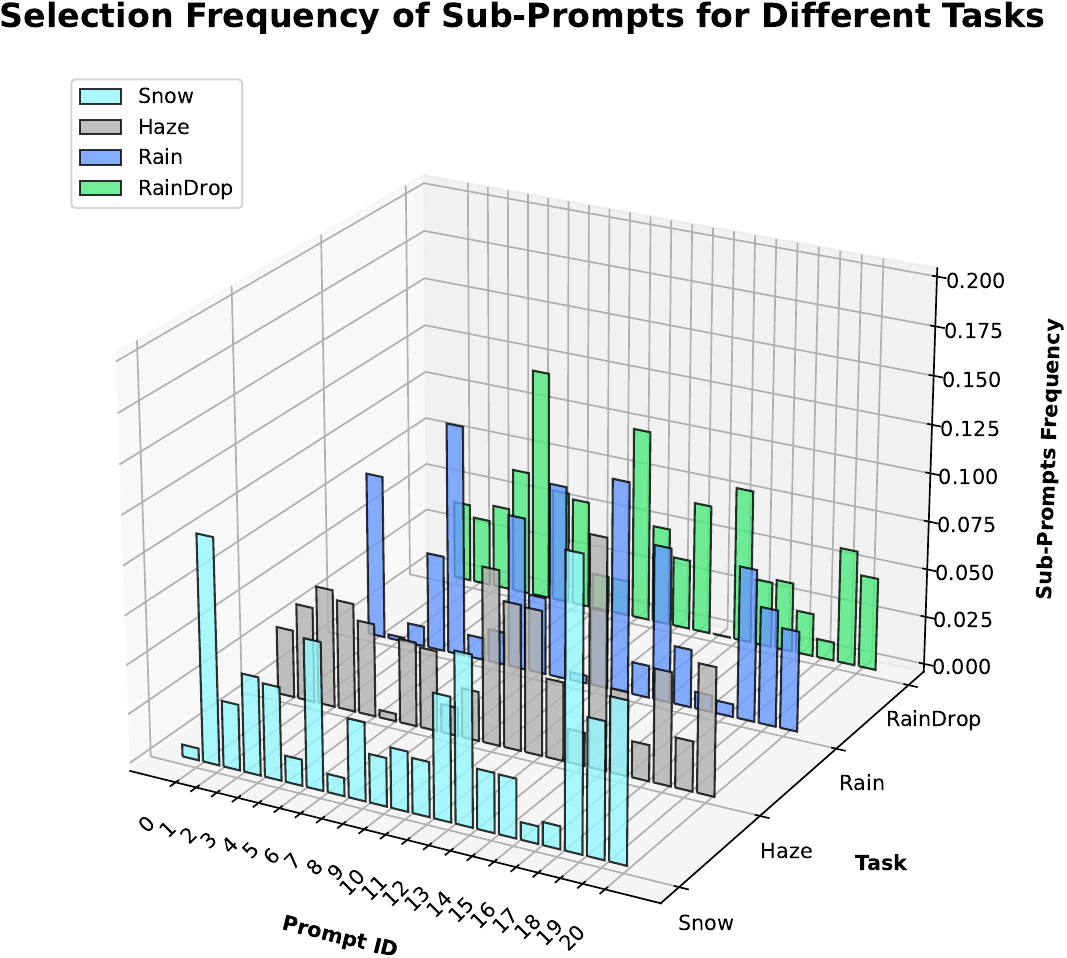}
    \caption{\small{\textbf{The selection frequency of sub-prompts.} Some similar selection frequencies reflect the network's ability to adaptively exploit common attributes in some similarity between tasks (e.g. rain and raindrop). At the same time, the unique prompt frequencies highlight the flexibility to adapt to the specific characteristics of each weather condition.}}
    \label{selection}
    \vspace{-0.3cm}
\end{figure}
\begin{equation}
\begin{aligned}
\boldsymbol{\mathcal{K}}^i_s \overset{\mathrm{match}}{\longleftrightarrow} \boldsymbol{\mathcal{P}}_{s}^{i}, \quad 
\boldsymbol{\mathcal{F}}_{e} \in &\mathbb{R}^{H\times W\times D} \overset{\mathrm{mean}}{\longrightarrow} \boldsymbol{\mathcal{F}}_{e}^{mean} \in \mathbb{R}^{1\times D}, \\
&\delta(\boldsymbol{\mathcal{K}}^i_s,\boldsymbol{\mathcal{F}}_{e}),
\end{aligned}
\end{equation}
where $\delta(\cdot)$ denotes the similarity calculation (we empirically choose cosine similarity). We employ this metric to let the network select the most appropriate sub-prompts from the pool to form effective weather-prompts.
\begin{equation}
\boldsymbol{\mathcal{P}}_{w} = \bigcup_{i=1}^{k} \boldsymbol{\mathcal{K}}_{s}^i \quad \text{if} \quad \delta(\boldsymbol{\mathcal{K}}^i_s, \boldsymbol{\mathcal{F}}_{e}) \geq \delta(\boldsymbol{\mathcal{K}}^{i+1}_s, \boldsymbol{\mathcal{F}}_{e}),
\end{equation}
where $k$ is the number of sub-prompts with the top-$k$ similarity we selected. $\bigcup_{i=1}^{k}$ denotes the concatenation of individual sub-prompts to construct the weather-prompts $\boldsymbol{\mathcal{P}}_{w}$, which embodies the most representative features of the input in relation to the given weather conditions. Such a manner can be understood as the network using sub-prompts to freely control the weather attributes that need to be learned (see Fig.\ref{selection}), which is novel and efficient compared to previous paradigms. Additionally, through t-SNE visualization of the weather-prompts under distinct weather scenarios, as illustrated in Fig.\ref{tsne_vis}, it is evident that our constructed weather-prompts not only preserve the unique attributes of each weather type but also leverage the commonalities and differences among them. This adaptive combination tailored to individual samples achieves exceptional performance, aligning with the concept of "teaching tailored to talent".

\noindent\textbf{\textcolor[RGB]{63, 167, 246}{Discussion I:}} 
Recently, prompt learning has been used in image restoration~\cite{potlapalli2023promptir}. Nonetheless, such an approach often relied on shared parameters to address various degradation scenarios, leading to potential interference among different degradations and overlooking the unique features of instance-wise degradations. We aim to implement a novel strategy "prompt pool". It enables the network to adaptively select appropriate sub-prompts in response to the specific degradation present in the input, thereby concentrating on the inter-attributes and intra-attributes of the weather degradations.
\vspace{-0.1cm}
\subsection{General Prompts for Scene Modeling} \label{sec:general prompts}
\vspace{-0.1cm}
\noindent{\textbf{Motivation II:}} Scene information provides guidance for the reconstruction of degraded residuals. In contrast to previous methods that solely focus on understanding degradations~\cite{valanarasu2022transweather,weatherdiff,zhu2023learning}, we claim that modeling the scene content is another critical factor in enhancing performance. Inspired by this insight, we contemplate whether clean backgrounds in degraded images possess distinguishable characteristics. Utilizing t-SNE for visualization, we observe the distribution among clean images in Fig.\ref{motivation}. There is often a significant distinction between degradations and background, while clean images share commonalities within the latent space. Consequently, we propose general prompts that encourage the network to boost representation with respect to the background.

The proposed general prompts $\boldsymbol{\mathcal{P}}_{g}\in \mathbb{R}^{L_{g}\times D}$, unlike the degradation-specific sub-prompts, are designed to encapsulate the common attributes of the scene across various weather distortions. It serves as a versatile anchor within the representational space, fostering a consistent perception of the background. Therefore, for the initialization of prompts, we seek to impose an explicit constraint that directs the learning towards the general attributes of the scenes. It will
ensure that the general prompts are not disturbed by the varying degrees of degradations.
\begin{figure}[t]
\vspace{-0.2cm}
  \centering
      \setlength{\abovecaptionskip}{0.2cm} 
    \setlength{\belowcaptionskip}{-0.2cm}
  \includegraphics[width=0.48\textwidth]{ 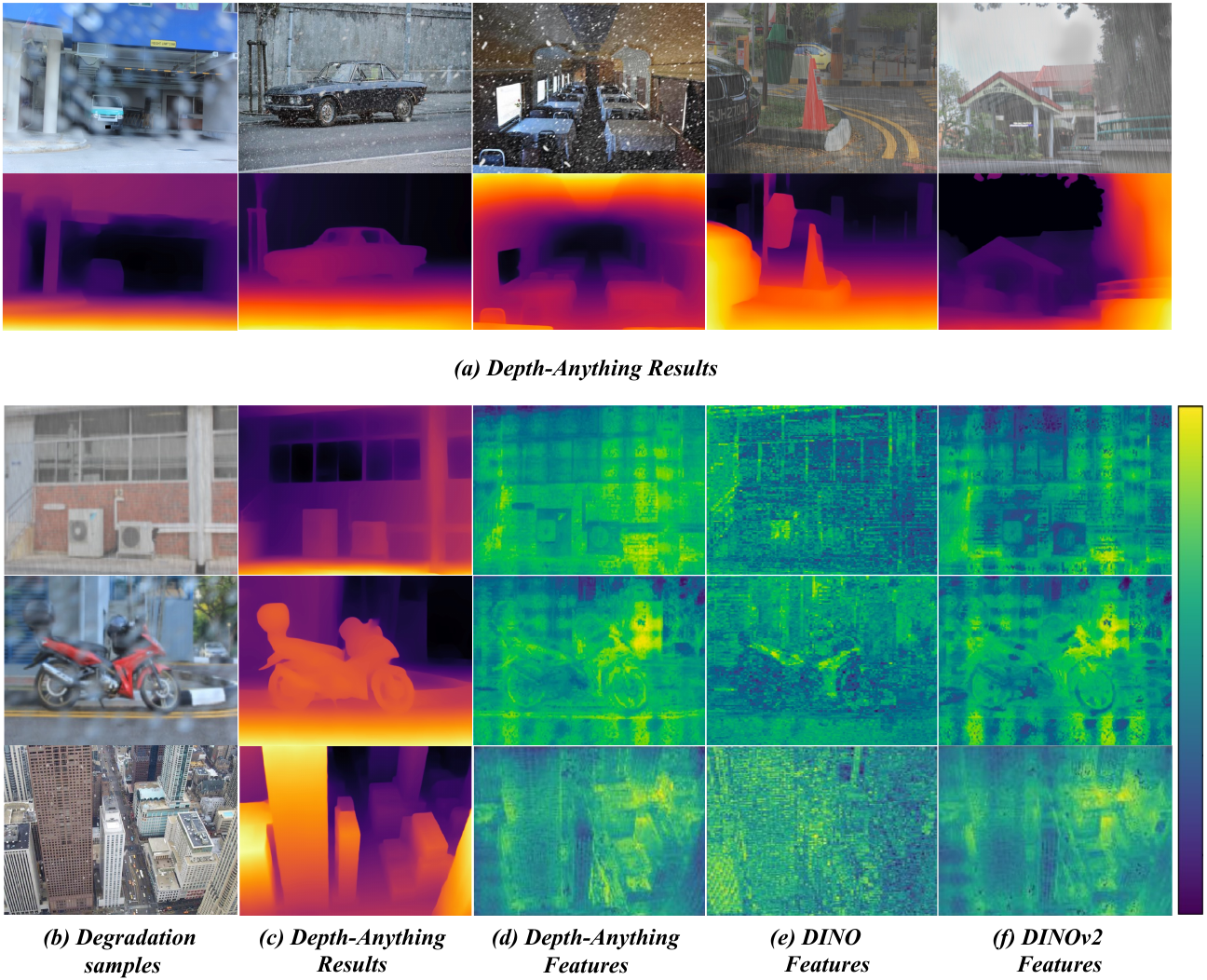}
  \caption{\textbf{Motivation of Depth-Anything~\cite{yang2024depth} as a constraint.} Depth-Anything has degradation-independent performance, and the intermediate features have better robustness than the previous pre-trained network~\cite{dino,oquab2023dinov2}.}
  \label{motivation_depth}
  \vspace{-0.3cm}
\end{figure}

\noindent{\textbf{Observation:}} Inspired by scene understanding~\cite{Chen_2019_CVPR,jiang2018self}, utilizing depth information has proven to effectively represent clean scenes. Additionally, adverse weather conditions is notably more susceptible to depth-related distortions compared to other degradations. Recently, the Depth-Anything~\cite{yang2024depth} model leverages extensive datasets and the robust representational capabilities of DINOv2~\cite{oquab2023dinov2} to develop a depth estimation model that applies to any scene.  As illustrated in

\noindent  
 Fig.\ref{motivation_depth} (a), we observe that depth maps estimated by Depth-Anything are almost unaffected by degradations, which to some extent demonstrates the robustness of the intermediate hidden features in scene representation (Fig.\ref{motivation_depth} (b)). 
Motivated by this discovery, we claim that the latent space features can be used to impose an explicit constraint on the general prompts, thereby better focusing on the background portion. 

\noindent{\textbf{Depth-Anything Constraint:}} 
To direct the general prompts $\boldsymbol{\mathcal{P}}_{g} \in \mathbb{R}^{L_g \times D}$ towards a more nuanced perceiving of the background, we integrate the latent features from the Depth-Anything model using an attention-based mechanism. Specifically, we define a cross-attention operation where the general prompts form the queries, and the keys and values are derived from the Depth-Anything features. 
Let $\boldsymbol{\mathcal{F}}_{{d}} \in \mathbb{R}^{H \times W \times D}$ represents the depth-aware features, where $H$ and $W$ are the dimensions of the feature map. The cross-attention mechanism is then given by:
\begin{equation}
\boldsymbol{\mathcal{P}}_{gd} = \text{softmax}\left(\frac{\boldsymbol{\mathcal{Q}}_{g} \boldsymbol{\mathcal{K}}_{{d}}^T}{\sqrt{\boldsymbol{\mathcal{D}}}}\right) \boldsymbol{\mathcal{V}}_{{d}},
\end{equation}
Obtained general prompts $\boldsymbol{\mathcal{P}}_{gd}$ adaptively integrate scene information, providing sufficient prior knowledge for the subsequent perception of the background while mitigating the impact of degradations.

\vspace{-0.1cm}
\subsection{Contrastive Prompt Loss}\label{sec:prompt loss}
\vspace{-0.1cm}
The sections above illustrate two types of prompts as the condition for the diffusion model. Additionally, we introduce contrastive prompt loss. It aims to differentially enhance the representations of two uniquely designed prompts. These prompts act as conditions for the diffusion model, with one tailored to model weather degradations and the other, guided by the Depth-Anything model, to perceive the background information. Given the inherently different design objectives of each prompt type, they are hypothesized to act as negative samples for each other. For the positive samples of prompt type, we employ cosine similarity to draw them nearer to the constraint within the latent space.
The contrastive prompt loss \( \mathcal{L}_{{cp}} \) is defined as:
\begin{equation}\label{eq:cp}
\mathcal{L}_{{cp}} = \frac{1}{b} \frac{1}{k} \sum_{j=1}^{b} \sum_{i=1}^{k} \left[ \gamma\left( \boldsymbol{\mathcal{K}}_{gd}, \boldsymbol{\mathcal{F}}_{{d}}^{mean} \right) - \gamma\left( \boldsymbol{\mathcal{K}}_{s}^{i}, \boldsymbol{\mathcal{K}}_{gd} \right) \right],
\end{equation}
where $b$ is the batch size. $\boldsymbol{\mathcal{K}}_{gd}$ represents the learnable key matched for general prompts, $\gamma(\cdot)$ denotes the $1-\delta(\cdot)$, which ensures the optimization process.

\noindent\textbf{\textcolor[RGB]{130, 234, 80}{Discussion II:}} While our approach draws inspiration from prior contrastive learning paradigms~\cite{wu2021contrastive,chen2022learning,Zheng_2023_CVPR}, it is distinctly different. 1). We do not require the construction of additional negative samples, as the two types of motivation-driven prompts within our design naturally serve as negatives for each other. 2). Prompts has explicit constraints that draw them closer to feature embeddings, eliminating the need for ground truth images as positive samples. 3). Our prompts can interact with high-dimensional features within the network directly, obviating the process for the traditional contrastive learning step of mapping via a pre-trained network~\cite{simonyan2014very} to a feature space.

\vspace{-0.1cm}
\subsection{Loss Function}
\vspace{-0.1cm}
To supervise our T$^{3}$-DiffWeather model, we employ the noise estimation loss Eq.\ref{eq:res} and the contrastive prompt loss Eq.\ref{eq:cp} during the noise estimation phase. Additionally, our contrastive prompt loss is designed to optimize the prompts adapted to different instance samples. Hence, during the training, we conduct the sampling process for restoring clean images and impose additional supervision on this process through reconstruction loss and contrastive prompt loss. This approach better constrains the optimization trajectory~\cite{jiang2023low,hou2024global} of the diffusion process and releases the potential of prompts. The overall objective function can be expressed as follows:
\begin{equation}
\begin{aligned}
   \mathcal{L}_{total} =  \lambda_{1}\mathcal{L}_{res}  + \lambda_{2}&\mathcal{L}_{{cp}} +  \lambda_{3}\left\|{(\boldsymbol{r}_{d}^{sample}+\boldsymbol{y})}-\boldsymbol{x}\right\|_\text{psnr} \\ +\lambda_{4}\mathcal{L}_{cp}^{sample}
\end{aligned}
\end{equation}
where $\text{psnr}$ denotes the PSNR loss~\cite{chen2021hinet,chen2022dual} we choose empirically. $\lambda_{1}$, $\lambda_{2}$, $\lambda_{3}$ and $\lambda_{4}$ are set to 1 empirically.

\vspace{-0.1cm}
\section{Experiments}
\vspace{-0.1cm}

\begin{figure*}[t]
\parbox{.41\textwidth}{
\centering
\setlength{\abovecaptionskip}{0.1cm} 
\setlength{\belowcaptionskip}{-0.0cm}
\captionsetup{font=small}
\captionof{table}{{\underline{\textbf{Desnowing.}}}}\label{desnow}
\scalebox{0.665}{
\renewcommand\arraystretch{1.1}
\begin{tabular}{l c c c c  }
\hline
\rowcolor{mygray}
& \multicolumn{2}{c}{Snow100K-S~\cite{liu2018desnownet}} & \multicolumn{2}{c}{Snow100K-L~\cite{liu2018desnownet}} \\
\cline{2-3}\cline{4-5}
\rowcolor{mygray}
\multirow{-2}*{Method} & PSNR $\uparrow$ & SSIM $\uparrow$  & PSNR $\uparrow$ & SSIM $\uparrow$ \\
\hline
\hline
SPANet\textcolor{gray_venue}{\scriptsize{[CVPR'19]}}~\cite{wang2019spatial} & 29.92 & 0.8260 &  23.70 & 0.7930 \\
JSTASR\textcolor{gray_venue}{\scriptsize{[ECCV'20]}}~\cite{chen2020jstasr} & 31.40 & 0.9012 & 25.32 & 0.8076 \\
RESCAN\textcolor{gray_venue}{\scriptsize{[ECCV'18]}}~\cite{li2018recurrent} & 31.51 & 0.9032 & 26.08 & 0.8108 \\
DesnowNet\textcolor{gray_venue}{\scriptsize{[TIP'18]}}~\cite{liu2018desnownet} & 32.33 & 0.9500 &  27.17 & 0.8983 \\
DDMSNet\textcolor{gray_venue}{\scriptsize{[TIP'21]}}~\cite{zhang2021deep} & 34.34 & 0.9445 & 28.85 & 0.8772 \\
MPRNet\textcolor{gray_venue}{\scriptsize{[CVPR'21]}}~\cite{mpr}  &{34.97} & {0.9457} & {29.76} &{0.8949} \\
NAFNet\textcolor{gray_venue}{\scriptsize{[ECCV'22]}}~\cite{chen2022simple}  &{34.79} & {0.9497} &{30.06} &{0.9017} \\
Restormer\textcolor{gray_venue}{\scriptsize{[CVPR'22]}}~\cite{zamir2021restormer} &\textbf{35.03}$^{\triangle}$ & \textbf{0.9487}$^{\triangle}$ &\textbf{30.52}$^{\triangle}$ &\textbf{0.9092}$^{\triangle}$ \\
\midrule
\midrule
All-in-One\textcolor{gray_venue}{\scriptsize{[CVPR'20]}}~\cite{allinone} & - & -  & 28.33 & 0.8820 \\
TransWeather\textcolor{gray_venue}{\scriptsize{[CVPR'22]}}~\cite{valanarasu2022transweather} & 32.51 & 0.9341 & 29.31 & 0.8879 \\
TKL$\&$MR\textcolor{gray_venue}{\scriptsize{[CVPR'22]}}~\cite{chen2022learning} &34.80 &0.9483 &30.24 &0.9020\\ 
{WeatherDiff$_{64}$}\textcolor{gray_venue}{\scriptsize{[PAMI'23]}}\cite{weatherdiff} & {35.83} & {0.9566} & 
{30.09} & {0.9041} \\
{WeatherDiff$_{128}$}\textcolor{gray_venue}{\scriptsize{[PAMI'23]}}\cite{weatherdiff} & {35.02} & {0.9516} & {29.58} & {0.8941} \\
{AWRCP}\textcolor{gray_venue}{\scriptsize{[ICCV'23]}}\cite{ye2023adverse} & \underline{36.92} & \underline{0.9652} & \underline{31.92} & \underline{0.9341} \\
\midrule
\rowcolor{mygray}
\scalebox{1.5}{{$\star$}} {T$^{3}$-DiffWeather (Ours)} & \textbf{37.51}  & \textbf{0.9664} &  \textbf{32.37} & \textbf{0.9355}   \\

\bottomrule
\end{tabular}

}}
\vspace{-0.1cm}
\hfill
\parbox{.29\textwidth}{
\centering
\setlength{\abovecaptionskip}{0.1cm} 
\setlength{\belowcaptionskip}{-0cm}
\captionsetup{font=small}
\captionof{table}{{\underline{\textbf{Deraining \& Dehazing. }}}}\label{derainhaze}
\scalebox{0.661}{
\renewcommand\arraystretch{1.1}
\begin{tabular}{l c c }
\hline 
\rowcolor{mygray}
& \multicolumn{2}{c}{Outdoor-Rain~\cite{Li_2019_CVPR}}\\
\cline{2-3}
\rowcolor{mygray}
\multirow{-2}*{Method}& PSNR $\uparrow$ & SSIM $\uparrow$ \\
\hline
\hline
\specialrule{0em}{7pt}{8pt} 
CycleGAN\textcolor{gray_venue}{\scriptsize{[ICCV'17]}}~\cite{zhu2017unpaired} & 17.62 & 0.6560 \\
pix2pix\textcolor{gray_venue}{\scriptsize{[ICCV'17]}}~\cite{pix2pix} & 19.09 & 0.7100 \\
HRGAN\textcolor{gray_venue}{\scriptsize{[CVPR'19]}}~\cite{li2019heavy} & 21.56 & 0.8550 \\
PCNet\textcolor{gray_venue}{\scriptsize{[TIP'21]}}~\cite{jiang2021rain} & 26.19 & 0.9015 \\
MPRNet\textcolor{gray_venue}{\scriptsize{[CVPR'21]}}~\cite{mpr} & {28.03} & {0.9192} \\ 
NAFNet\textcolor{gray_venue}{\scriptsize{[ECCV'22]}}~\cite{chen2022simple} & {29.59} &{0.9027}
\\
Restormer\textcolor{gray_venue}{\scriptsize{[CVPR'22]}}~\cite{zamir2021restormer} &\textbf{29.97}$^{\triangle}$ &\textbf{0.9215}$^{\triangle}$ \\
\midrule
\midrule
All-in-One\textcolor{gray_venue}{\scriptsize{[CVPR'20]}}~\cite{allinone} & 24.71 & 0.8980 \\
TransWeather\textcolor{gray_venue}{\scriptsize{[CVPR'22]}}~\cite{valanarasu2022transweather} & 28.83 & 0.9000 \\

TKL$\&$MR\textcolor{gray_venue}{\scriptsize{[CVPR'22]}}~\cite{chen2022learning} &{29.92} &{0.9167} \\

{WeatherDiff$_{64}$}\textcolor{gray_venue}{\scriptsize{[PAMI'23]}}~\cite{weatherdiff} & {29.64} & {0.9312}  \\
{WeatherDiff$_{128}$}\textcolor{gray_venue}{\scriptsize{[PAMI'23]}}~\cite{weatherdiff} & {29.72} & {0.9216} \\
{AWRCP}\textcolor{gray_venue}{\scriptsize{[ICCV'23]}}\cite{ye2023adverse} & \underline{31.39} & \underline{0.9329}\\

\midrule
\rowcolor{mygray}
\scalebox{1.5}{{$\star$}} {T$^{3}$-DiffWeather (Ours)} & \textbf{31.99} & \textbf{0.9365} \\
\bottomrule

\end{tabular}}

}
\vspace{-0.1cm}
\hfill
\parbox{.290\textwidth}{
\centering
\setlength{\abovecaptionskip}{0.1cm} 
\setlength{\belowcaptionskip}{-0cm}
\captionsetup{font=small}
\captionof{table}{{\underline{\textbf{Raindrop Removal.}}}}\label{raindrop}
\scalebox{0.661}{
\renewcommand\arraystretch{1.1}
\begin{tabular}{l c c }
\hline 
\rowcolor{mygray}
& \multicolumn{2}{c}{RainDrop~\cite{qian2018attengan}}\\
\cline{2-3}
\rowcolor{mygray}
\multirow{-2}*{Method}& PSNR $\uparrow$ & SSIM $\uparrow$ \\
\hline
\hline
\specialrule{0em}{7pt}{8pt} 
pix2pix\textcolor{gray_venue}{\scriptsize{[ICCV'17]}}~\cite{pix2pix} & 28.02 & 0.8547 \\
DuRN\textcolor{gray_venue}{\scriptsize{[CVPR'19]}}~\cite{liu2019dual} & 31.24 & 0.9259 \\ 
RaindropAttn\textcolor{gray_venue}{\scriptsize{[ICCV'19]}}~\cite{quan2019deep} & 31.44 & 0.9263  \\
AttentiveGAN\textcolor{gray_venue}{\scriptsize{[CVPR'18]}}~\cite{qian2018attengan} & 31.59 & 0.9170  \\
CCN\textcolor{gray_venue}{\scriptsize{[CVPR'21]}}~\cite{quaninonego} & {31.34} & {0.9286}  \\
IDT\textcolor{gray_venue}{\scriptsize{[PAMI'22]}}~\cite{IDT} & {31.87} & {0.9313}  \\
UDR-S$^{2}$Former\textcolor{gray_venue}{\scriptsize{[ICCV'23]}}~\cite{chen2023sparse} & \textbf{32.64}$^{\triangle}$ & \textbf{0.9427}$^{\triangle}$ \\
\midrule
\midrule
All-in-One\textcolor{gray_venue}{\scriptsize{[CVPR'20]}}~\cite{allinone} & {31.12} & {0.9268} \\
TransWeather\textcolor{gray_venue}{\scriptsize{[CVPR'22]}}~\cite{valanarasu2022transweather} & 30.17 & 0.9157 \\
TKL$\&$MR\textcolor{gray_venue}{\scriptsize{[CVPR'22]}}~\cite{chen2022learning} &{30.99} &{0.9274} \\
{WeatherDiff$_{64}$}\textcolor{gray_venue}{\scriptsize{[PAMI'23]}}~\cite{weatherdiff} & {30.71} & {0.9312}  \\ 
{WeatherDiff$_{128}$}\textcolor{gray_venue}{\scriptsize{[PAMI'23]}}~\cite{weatherdiff} & 29.66 & 0.9225 \\
{AWRCP}\textcolor{gray_venue}{\scriptsize{[ICCV'23]}}\cite{ye2023adverse} & \underline{31.93} & \underline{0.9314} \\

\midrule
\rowcolor{mygray}
\scalebox{1.5}{{$\star$}} {T$^{3}$-DiffWeather (Ours)} & \textbf{32.66} & \textbf{0.9411}  \\

\bottomrule
\end{tabular}}

}
\caption{These tables provide quantitative comparisons with state-of-the-art image desnowing, deraining, and adverse weather removal methods, employing PSNR and SSIM as metrics—where higher values signify better restoration. The best and second-best metrics are shown with \textbf{bold text} and \underline{underlined text}, respectively. The triangle $\triangle$ represents the SOTA metric trained on a single dataset. Above half of the tables present comparisons of task-specific methods for a single dataset, while the bottom section showcases the performance of the proposed T$^{3}$-DiffWeather method across all four test sets against state-of-the-art adverse weather solutions, including All-in-One~\cite{allinone}, TransWeather~\cite{valanarasu2022transweather}, TKL\&MR~\cite{chen2022learning}, WeatherDiffusion~\cite{weatherdiff} and AWRCP~\cite{ye2023adverse}.}
\label{fig:three_tables}
\vspace{-0.3cm}
\end{figure*}

\vspace{-0.1cm}
\subsection{Implementation}
\vspace{-0.1cm}
\begin{figure*}[!t]
    \centering
    \setlength{\abovecaptionskip}{0.2cm} 
    \setlength{\belowcaptionskip}{-0.2cm}
    \includegraphics[width=17.5cm]{ 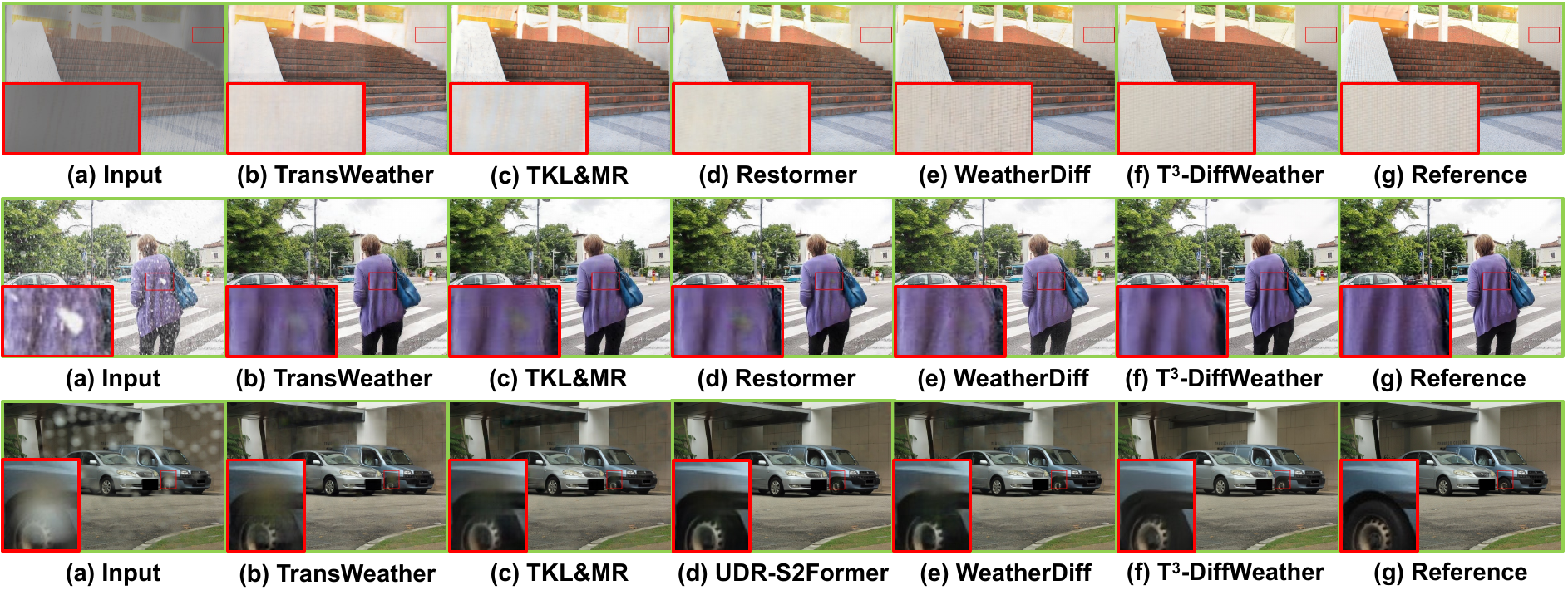}
    \caption{Visual comparisons in adverse weather conditions on Snow100K~\cite{liu2018desnownet}, Outdoor-Rain~\cite{valanarasu2022transweather} and RainDrop~\cite{qian2018attengan} datasets.}
    \label{com_syn1}
    \vspace{-0.4cm}
\end{figure*}
\noindent{\textbf{Pipeline implementation.}}
T$^{3}$-DiffWeather builds upon the backbone followed by previous diffusion design~\cite{luo2023refusion}. We employ uniform initialization techniques to set up the weights for sub-prompts in the prompt pool and general prompts, including their respective keys. Specifically, we designate a total of $20$ sub-prompts within the prompt pool, with each sub-prompt comprising $64$ tokens ($L_{s}$), from which we select the top $5$ ($k$) to form the required weather-prompts. The token number for general prompts ($L_{g}$) is set at $256$ to ensure the balance between performance and manageable overhead. When constraining the general prompts using Depth-Anything~\cite{yang2024depth}, we utilize the ViT-S architecture, which demands minimal memory usage while maintaining robustness. During the diffusion process, we opt for DDIM~\cite{ddim} sampling. Owing to our focus on reconstructing degradation residuals and the rich representations of the condition, setting the number of sampling steps to merely $2$ suffices to achieve impressive performance. Additional architectural details can be found in the supplementary materials.
\begin{figure*}[!t]
    \centering
    \setlength{\abovecaptionskip}{0.2cm} 
    \setlength{\belowcaptionskip}{-0.2cm}
    \includegraphics[width=17.5cm]{ 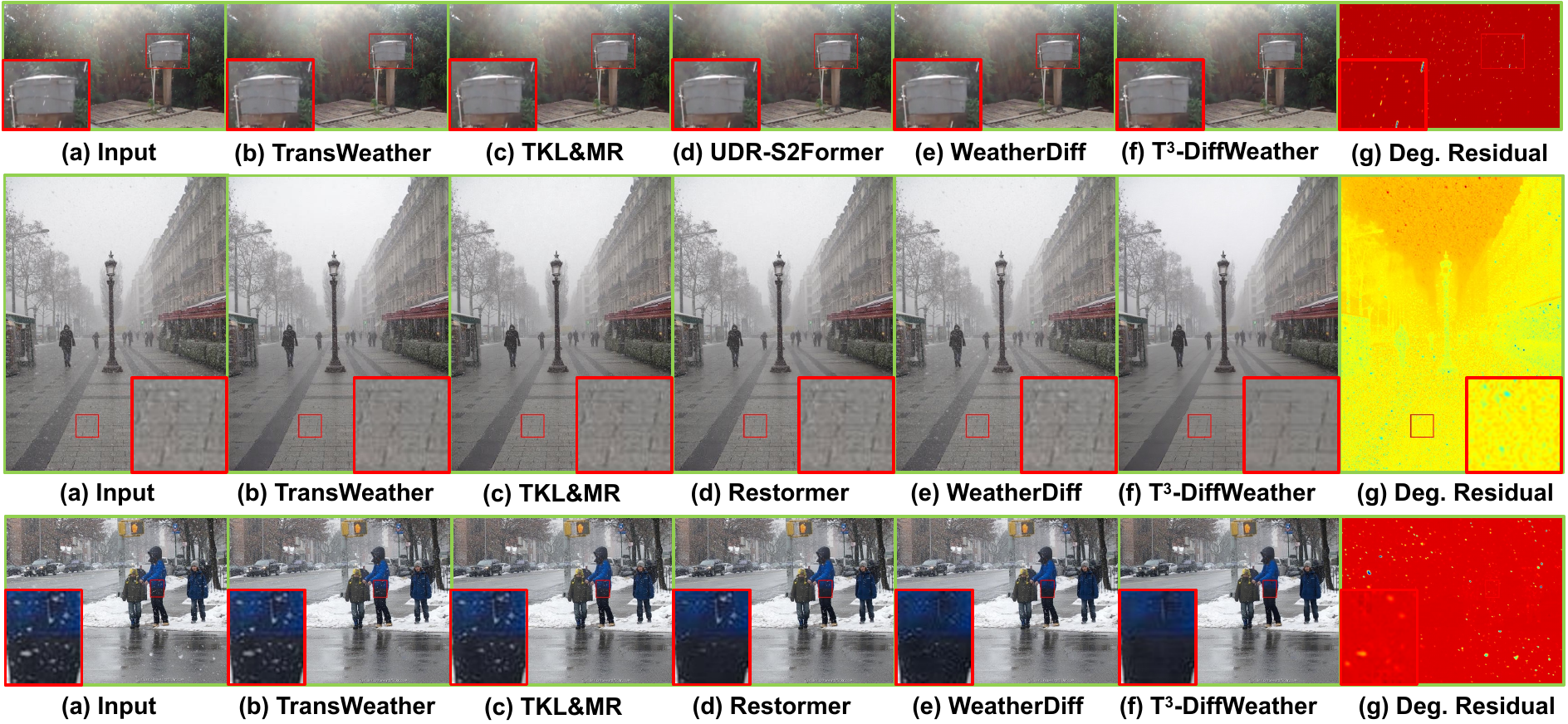}
    \caption{Visual comparisons of the real-world adverse weather samples.}
    \label{com_real2}
\end{figure*}

\noindent{\textbf{Training details.}}
To train our T$^{3}$-DiffWeather model, we leverage the comprehensive AllWeather dataset referenced in~\cite{valanarasu2022transweather}, including $18,069$ images from the Snow100K~\cite{liu2018desnownet}, Outdoor-Rain~\cite{Li_2019_CVPR}, and RainDrop~\cite{qian2018attengan} datasets, the same as previous adverse weather restoration methods~\cite{allinone,valanarasu2022transweather,weatherdiff,ye2023adverse}. 
Our T$^{3}$-DiffWeather pipeline is developed on the PyTorch framework and undergoes training on two NVIDIA A800 GPUs. This pipeline includes $800$K training iterations, utilizing the Adam optimizer with momentum parameters set to $0.9$ and $0.995$. Training commences with an initial learning rate of $1.5\times10^{-4}$, which is reduced using a cosine annealing schedule. To promote stability during the learning phase, an exponential moving average strategy weighted at $0.995$ is employed for parameter updates, as supported by findings in \cite{nichol2021improved} and \cite{song2020improved}. The diffusion procedure consists of $1,000$ timesteps, labeled as $T$, with an incrementally ascending noise schedule $\beta_{t}$ ranging from $0.0001$ to $0.02$. The training employs image patches of  $256\times256$ pixels. Augmentation techniques like horizontal flipping and fixed-angle random rotation are used in the training. Please refer to the supplementary material to view detailed training and testing dataset configurations.

\vspace{-0.1cm}
\subsection{Quantitative comparison}
\vspace{-0.1cm}
We perform a comparative analysis of metrics between synthetic and real datasets. Specifically, we compare the model performance for a single task and the performance of a multi-weather image restoration model trained on multiple weather datasets. Our quantitative analysis reveals the competitive advantage of our T$^{3}$-DiffWeather pipeline over existing state-of-the-art algorithms in image restoration with various weather impacts. As shown in Tab.\ref{desnow}, T$^{3}$-DiffWeather achieves excellent performance in image snow removal, as evidenced by the highest PSNR and SSIM metrics on the Snow100K-S~\cite{liu2018desnownet} and Snow100K-L~\cite{liu2018desnownet} datasets. It is particularly noteworthy that the PSNR on Snow100K-S is 1.68db higher than the previous best diffusion model, WeatherDiffusion~\cite{weatherdiff}, indicating a significant improvement in recovery quality. This is mainly due to our new pipeline design and suitable and effective conditions. In addition, our method ranks first in the deraining and dehazing task (Tab.\ref{derainhaze}) and maintains the leading position in the raindrop removal (Tab.\ref{raindrop}).

\vspace{-0.1cm}
\subsection{Visual Comparison}
\vspace{-0.1cm}
Fig.\ref{com_syn1} visually compares state-of-the-art image restoration techniques on a synthetic dataset designed to simulate real-world conditions. WeatherDiffusion~\cite{weatherdiff} marginally enhances detail definition but does not remove degradations entirely in some areas. Restormer improves color fidelity but does not entirely eliminate synthetic artifacts. T$^{3}$-DiffWeather, our method, markedly improves texture and color accuracy, closely matching the reference. It significantly reduces synthetic distortions, maintaining scene authenticity, as seen in the detailed insets.

Furthermore, Fig.\ref{com_real2} also shows a visual comparison of restoration methods applied to images of real-world scenarios. Also based on a diffusion model, our method can better remove degradation in the real world and restore complex scene textures than WeatherDiffusion~\cite{weatherdiff}.
In addition, UDR-S2Former~\cite{chen2023sparse} has deficiencies in handling real rainy scenes. In comparison, our method visually removed all degradations details, which proves the competitiveness of our method compared to specific methods. We also show the heat map of our degradation residual. We find that our method always focuses on degradations in terms of the restoraiton object, which proves the effectiveness of our pipeline.

\vspace{-0.1cm}
\section{Evaluating Real-World Performance}\label{sec:performance-real}
\vspace{-0.1cm}
To further demonstrate the superiority of our paradigm in the real world, we conduct a quantitative comparison of real-world datasets. Tab.\ref{realraindsresults} presents an evaluation of our T$^{3}$-DiffWeather approach on datasets captured under real-world meteorological conditions~\cite{chen2023sparse,quaninonego,ba2022not}. Our T$^{3}$-DiffWeather method achieves impressive scores, emphasizing its excellent adaptability to different environmental conditions. This shows that our "teaching tailored to talent" paradigm can allow the network to adaptively utilize the required attributes, which is effective for diverse real-life scenarios. It shows that our method is widely applicable to actual environments.

\begin{table}[t]
\vspace{-0.3cm}
\centering
\setlength{\abovecaptionskip}{0.0cm} 
\setlength{\belowcaptionskip}{-0.0cm}
\captionof{table}{{{{{Com. on more datasets.}}}}}
\label{realraindsresults}
\resizebox{6.8cm}{!}{
\setlength\tabcolsep{2pt}
\renewcommand\arraystretch{1.1}
\begin{tabular}{c|cc|cc}
\hline
\rowcolor{mygray}
  &  \multicolumn{2}{c|}{ \textbf{RainDS-Real(RDS)} }  & \multicolumn{2}{c}{ \textbf{GT-RAIN} } \\\cline{2-5}
 \rowcolor{mygray}
\multirow{-2}*{Method}  & PSNR $\uparrow$ & SSIM $\uparrow$  & PSNR $\uparrow$ & SSIM $\uparrow$  \\\hline\hline
WGWS-Net~\cite{zhu2023learning} &{{20.79}}&{{0.603}}&20.65&0.608 \\
WeatherDiff$_{128}$~\cite{weatherdiff} &{{21.09}}&{{0.605}} &20.83&0.613\\
AWRCP~\cite{ye2023adverse} &{{21.31}}&{{0.607}}&20.97&0.615 \\
\midrule
\rowcolor{mygray}
\scalebox{1.5}{{$\star$}} T$^{3}$-DiffWeather (ours) &{\textbf{21.96}}&{\textbf{0.612}} &\textbf{21.27}&\textbf{0.619}\\
\bottomrule
\end{tabular}
        }
        \vspace{-0.2cm}
\end{table}

\vspace{-0.1cm}
\subsection{Comparison of Parameters and Complexity}
\vspace{-0.1cm}
\begin{table}[h]
\vspace{-0.3cm}
  \centering
\setlength{\abovecaptionskip}{0.0cm} 
\setlength{\belowcaptionskip}{-0.0cm}
  \caption{Com. of parameters and GFLOPs (256$\times$256 resolution) for diffusion process and the regressive model.}
  \resizebox{7cm}{!}{
  \begin{tabular}{lcc}
    \hline 
        \rowcolor{mygray}
    \textbf{Method} & \textbf{\#Params} & \textbf{\#GFLOPs} \\
    \hline
    \multicolumn{3}{c}{\textbf{Image Restoration (Regression)}} \\
    Restormer~\cite{zamir2021restormer} & 25.3M & 140.92G \\
    \hline
    \multicolumn{3}{c}{\textbf{Image Restoration (Diff)}} \\
    IR-SDE~\cite{irsde} & 135.3M & 119.1G$\times$100 steps \\
    Refusion~\cite{luo2023refusion} & 131.4M & 63.4G$\times$50 steps \\
    \hline 
    \multicolumn{3}{c}{\textbf{Adverse Weather Restoration (Diff)}} \\
    WeatherDiffusion~\cite{weatherdiff} & 113.68M & 248.4G$\times$25 steps \\
    T$^{3}$-DiffWeather (ours) & 69.38M & 59.82G$\times$2 steps \\
    \hline 
  \end{tabular}}
  \vspace{-0.2cm}
  \label{tab:para_comparison}
\end{table}

As shown in Tab.\ref{tab:para_comparison}, our pipeline significantly reduces the number of parameters required for diffusion compared to previous designs, while decreasing the GFLOPs compared with the regressive model~\cite{zamir2021restormer}. Moreover, with only two steps in the sampling process, the computational complexity at a 256$\times$256 resolution is a mere \textbf{1/52} of that of the SOTA WeatherDiffusion~\cite{weatherdiff}. Additionally, the computational complexity of the single image restoration diffusion architecture Refusion~\cite{luo2023refusion} is nine times of ours, underscoring the efficacy of proposed conditions and constraints and the superiority of our holistic approach aimed at reconstructing degradation residuals.

\vspace{-0.1cm}
\subsection{Ablation Studies}
\vspace{-0.1cm}
In order to verify the efficacy of each key component of T$^{3}$-DiffWeather, we conduct a series of ablation experiments. All these variants are trained using the same configurations as in the implementation details, and the ablation results are tested on Outdoor-Rain~\cite{Li_2019_CVPR}.

\begin{wraptable}{l}{4.2cm}
\vspace{-0.5cm}
\centering
\setlength{\abovecaptionskip}{0cm} 
\setlength{\belowcaptionskip}{0cm}
\caption{Abl. of Prompt Pool.}
\label{ab:prompt pool}
\resizebox{4.2cm}{!}{
\renewcommand\arraystretch{1.1}
\begin{tabular}{l||cc}
\hline 
\rowcolor{mygray}
\textbf{Method} &  \textbf{PSNR} $\uparrow$ & \textbf{SSIM} $\uparrow$ \\
\hline
\hline
w/o. prompt pool  & 31.05 & 0.9325 \\
w/o. matched keys  & 31.72 & 0.9349 \\
w. prompt~\cite{potlapalli2023promptir} & 31.38 & 0.9330 \\
w. prompt pool (Ours)  & \textbf{31.99} & \textbf{0.9365} \\
\hline 
\rowcolor{mygray}
\textbf{Length} $L_{s}$&  \textbf{PSNR} $\uparrow$ & \textbf{SSIM} $\uparrow$ \\
\hline
\hline
32  (Sub-prompts)  & {31.79} & 0.9358 \\
64  (Sub-prompts) (Ours) & {31.99} & {0.9365} \\
128 (Sub-prompts) & \textbf{32.04} & \textbf{0.9366} \\
\hline 
\end{tabular}}
\vspace{-0.4cm}
\end{wraptable}
\noindent{\textbf{Effectiveness of Prompt Pool}}. The  Tab.\ref{ab:prompt pool} highlight the vital role of the proposed prompt pool in adverse weather restoration. Leveraging the advances in prompt learning, our prompt pool method autonomously crafts tailored sub-prompts for each specific degradation scenario. This approach has shown substantial improvements over methods without a prompt pool or with unmatched keys and significantly surpasses the previous design~\cite{potlapalli2023promptir}. With our method, the network selects the most representative sub-prompts with 64 token numbers of sub-prompts to balance the complexity and performance, affirming the prompt pool's utility in capturing the attributes of complex weather patterns. 

\begin{figure}[h]
\vspace{-0.2cm}
    \centering
    \setlength{\abovecaptionskip}{0.2cm} 
    \setlength{\belowcaptionskip}{-0.2cm}
    \includegraphics[width=5.5cm]{ 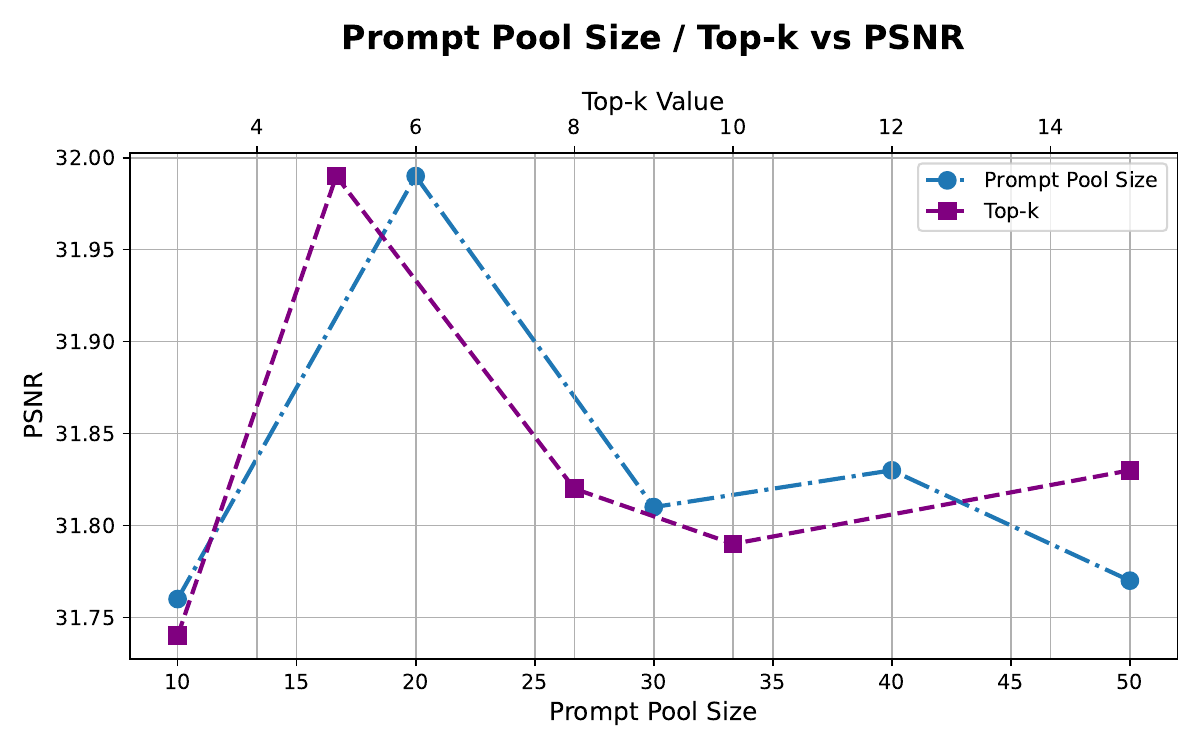}
    \caption{Abl. of Prompt pool size and $k$ value of top-$k$.}
    \label{tok_promptsize}
    \vspace{-0.2cm}
\end{figure}
\noindent{\textbf{Discussion of prompt pool size and top$\textbf{k}$}}.
Our results show that as the prompt pool size increases, the model has sufficient diversity in prompts to effectively capture and represent various degraded properties. Conversely, a prompt pool that is too large may introduce redundancy or noise, thereby compromising the model's resilience.
Additionally, for top-$k$, too fine-grained selection may lead to overfitting of poorly represented degradation attributes or ignoring valuable global information. Smaller top-$k$ values ensure that the most relevant sub-prompts are employed, resulting in more robust and general performance across various degraded images.

\begin{table}[h]
\vspace{-0.2cm}
\centering
\setlength{\abovecaptionskip}{0cm} 
\setlength{\belowcaptionskip}{0cm}
\caption{ Abl. of General Prompts.}
\label{ab:general prompts}
\resizebox{6cm}{!}{
\renewcommand\arraystretch{1.1}
\begin{tabular}{l||cc}
\hline 
\rowcolor{mygray}
\textbf{Method} &  \textbf{PSNR} $\uparrow$ & \textbf{SSIM} $\uparrow$ \\
\hline
\hline
w/o. General Prompts & 31.52 & 0.9342 \\
w/o. Depth-Anything~\cite{yang2024depth}  & 31.67 & 0.9349 \\
w. DINO~\cite{dino} & 31.77 & 0.9357 \\
w. DINOv2~\cite{oquab2023dinov2} & 31.82 & 0.9359 \\
w. Depth-Anything~\cite{yang2024depth} (Ours)  & \textbf{31.99} & \textbf{0.9365} \\
\hline 
\rowcolor{mygray}
\textbf{Length} $L_{g}$&  \textbf{PSNR} $\uparrow$ & \textbf{SSIM} $\uparrow$ \\
\hline
\hline
128  (General Prompts)  & 31.73 & 0.9354 \\
192 (General Prompts) & 31.88 & 0.9360 \\
256 (General Prompts) (Ours)& {31.99} & \textbf{0.9365} \\
320 (General Prompts) & \textbf{32.03} & 0.9365 \\
\hline 
\end{tabular}}
\vspace{-0.2cm}
\end{table}

\noindent{\textbf{Improvements of proposed General Prompts}}. Tab.\ref{ab:general prompts} illustrates the efficacy of incorporating the robust Depth-Anything~\cite{yang2024depth} features as a constraint for general prompts. The depth-aware features intrinsic to Depth-Anything have demonstrated superior performance over other pre-trained models (e.g. DINO~\cite{dino}, DINOv2~\cite{oquab2023dinov2}), particularly regarding scene understanding. Moreover, general prompts without such explicit constraints exhibit limitations in holistically background modeling. 

Furthermore, our experiments revealed a performance bottleneck associated with increasing the number of general prompts. To optimize efficiency without compromising gains, we determine that selecting 256 tokens yields the optimal balance, effectively avoiding unnecessary computational overhead.

\begin{table}[h]
\vspace{-0.2cm}
\centering
\setlength{\abovecaptionskip}{0cm} 
\setlength{\belowcaptionskip}{0cm}
\caption{ Com. of memory cost, PSNR, and SSIM across the different Depth-Anything~\cite{yang2024depth} architectures.}
\label{tab:memorycost}
\resizebox{7cm}{!}{
\renewcommand\arraystretch{1.1}
\begin{tabular}{l||c||cc}
\hline 
\rowcolor{mygray}
\textbf{Method} & \textbf{\#{Memory Cost}} & \textbf{PSNR} $\uparrow$ & \textbf{SSIM} $\uparrow$ \\
\hline
\hline
ViT-S-14 (Ours)& 115.22 MB & {31.99} & {0.9365} \\
ViT-B-14 & 402.29 MB & 31.96 & 0.9365 \\
ViT-L-14 & 1314.38 MB & \textbf{32.06} & \textbf{0.9366} \\
\hline 
\end{tabular}}
\vspace{-0.2cm}
\end{table}

\noindent{\textbf{Gains of different Depth-Anything Architecture}}. The ablation study shown in Tab.\ref{tab:memorycost} compares memory cost with image recovery quality of different Depth-Anything~\cite{yang2024depth} architectures. We found that the ViT-S-14 model stands out for its low memory consumption of only 115.22 MB, while still achieving competitive PSNR and SSIM metrics for our final results. In comparison, ViT-B-14 and ViT-L-14 require significantly more memory with little corresponding gain in PSNR or SSIM.

\begin{wraptable}{r}{3.8cm}
\vspace{-0.4cm}
\centering
\setlength{\abovecaptionskip}{0cm} 
\setlength{\belowcaptionskip}{0cm}
\caption{  Abl. of Contrastive Prompt Loss (CPL).}
\label{ab:cpl}
\resizebox{3.8cm}{!}{
\renewcommand\arraystretch{1.1}
\begin{tabular}{l||cc}
\hline 
\rowcolor{mygray}
\textbf{Method} &  \textbf{PSNR} $\uparrow$ & \textbf{SSIM} $\uparrow$ \\
\hline
\hline
w/o.  CPL & 31.71 & 0.9350 \\
w/o. Negative $\gamma$ & 31.81 & 0.9359 \\
w/o. Positive $\gamma$ & 31.77 & 0.9358 \\
w.   CPL (Ours)  & \textbf{31.99} & \textbf{0.9365} \\
\hline 
\end{tabular}}
\vspace{-0.4cm}
\end{wraptable}
\noindent{\textbf{Benefits of Contrast Prompt Loss (CPL). }} As shown in Tab.\ref{ab:cpl}, the CPL distinguishes designed prompts to guide the diffusion process. The table shows that the pull of explicit constraints for prompts and the push of two prompts based on different motivations play an important role in improving the learning effect. They improve the guidance performance of conditions on diffusion by promoting enhanced representation learning. More ablation experiments can be found in the supplementary material.

\vspace{-0.1cm}
\section{Conclusion}
\vspace{-0.1cm}
This paper draws inspiration from the prompt learning and the concept of "Teaching Tailored to Talent", proposing a novel T$^{3}$-DiffWeather. It utilizes weather-prompts constructed from free combinations of sub-prompts, and general prompts constrained by Depth-Anything to provide rich information for the diffusion process from the degradation and background perspectives. Experimental results demonstrate that our method achieves SOTA performance on various synthetic and real-world data sets, with excellent computational efficiency.

\section*{Acknowledgements}
This work is supported by the Guangzhou-HKUST(GZ) Joint Funding Program (No. 2023A03J0671), the National Natural Science Foundation of China (Grant No. 61902275), the Guangzhou Industrial Information and Intelligent Key Laboratory Project (No. 2024A03J0628), and Guangzhou-HKUST(GZ) Joint Funding Program (No. 2024A03J0618).

%
%
\bibliographystyle{ieeenat_fullname}
\bibliography{main}
\end{document}